\documentclass[pdflatex,sn-mathphys-num]{sn-jnl}

\usepackage{graphicx}%
\usepackage{multirow}%
\usepackage{amsmath,amssymb,amsfonts}%
\usepackage{amsthm}%
\usepackage{mathrsfs}%
\usepackage[title]{appendix}%
\usepackage{xcolor}%
\usepackage{textcomp}%
\usepackage{manyfoot}%
\usepackage{booktabs}%
\usepackage{algorithm}%
\usepackage{algorithmicx}%
\usepackage{algpseudocode}%
\usepackage{listings}%

\usepackage{booktabs}    
\usepackage{colortbl}     
\usepackage{xcolor}       

\usepackage{url}
\usepackage{multirow}

\usepackage{amsmath}    
\usepackage{amsfonts}   
\usepackage{amssymb}    
\usepackage{amsthm}     
\usepackage{graphicx}   
\usepackage{caption}    
\usepackage{subcaption}

\usepackage{algorithm}  
\usepackage{algpseudocode} 
\usepackage{enumitem}   
\usepackage{booktabs}

\DeclareMathOperator*{\argmax}{arg\,max}
\DeclareMathOperator*{\argmin}{arg\,min}

\newcommand{\h}{\mathbf{h}}
\newcommand{\w}{\mathbf{w}}

\newcommand{\W}{{\mathbf{W}}}

\theoremstyle{plain}
\newtheorem{theorem}{Theorem}[section]

\theoremstyle{definition}
\newtheorem{definition}[theorem]{Definition}

\theoremstyle{remark}

\begin{document}

\title[Hier-DETR]{Hierarchical Neural Collapse Detection Transformer for Class Incremental Object Detection}


\author[1]{\fnm{Duc Thanh} \sur{Pham}}


\author[1]{\fnm{Hong Dang} \sur{Nguyen}}
\author[1]{\fnm{Nhat Minh} \sur{Nguyen Quoc}}

\author[1]{\fnm{Linh} \sur{Ngo Van}}

\author[1]{\fnm{Sang} \sur{Dinh Viet}}


\author*[1]{\fnm{Duc Anh} \sur{Nguyen}} 


\affil*[1]{\orgdiv{School of Information and Communication Technology}, \orgname{Hanoi University of Science and Technology}, \orgaddress{\street{No.1 Dai Co Viet}, \city{Hanoi}, \postcode{100000}, \state{Hanoi}, \country{Vietnam}}}

%


\abstract{Recently, object detection models have witnessed notable performance improvements, particularly with transformer-based models. However, new objects frequently appear in the real world, requiring detection models to continually learn without suffering from catastrophic forgetting. Although Incremental Object Detection (IOD) has emerged to address this challenge, these existing models are still not practical due to their limited performance and prolonged inference time. In this paper, we introduce a novel framework for IOD, called \text{Hier-DETR}: \textbf{Hier}archical Neural Collapse \textbf{De}tection \textbf{Tr}ansformer, ensuring both efficiency and competitive performance. First, we propose a Hierarchical Neural Collapse Tree, inspired by the general orthogonal frame characteristics of the convergence of the class means, and the semantic hierarchical relationship between classes. After each query is labeled using the Hungarian Matching Algorithm, we pre-allocate new class classifiers of queries for each decoder's output layer as Hierarchical Neural Collapse Tree and fix this architecture as knowledge distillation while guiding each query to its class and super-class prototypes by Proxy-NCA. We conduct extensive experiments on COCO 2017, MTSD datasets and demonstrate that Hierarchical Neural Collapse helps model's queries more "stable" in Hungarian Matching, Hier-DETR achieves state-of-the-art performance in IOD settings: \text{48.4 AP} (70+10) in COCO and \text{45.46 AP} (150+71) in MSTD, outperforming with remarkable \text{+ 8 AP} and \text{+ 7.5 AP}  improvement compared to the previous best method.}

\maketitle

\section{Introduction}
\label{sec:intro}
In recent years, there exist numerous significant advancements in object detection (OD) \cite{rtdetr},\cite{codetr}. However, these models are designed to train from a fixed dataset based on the assumption that all objects’ class data are available at the training phase. Existing studies demonstrate that object detection frequently encounters continual learning scenarios in real-world implementations, especially in self-driving automobiles where novel traffic signs are regularly introduced. In light of this, incremental object detection such as: \cite{cldetr},\cite{sddgr} has been proposed to tackle this problem.

Currently, state-of-the-art detection models are the family of transformer-based detection models: DETR \cite{detr}, Deformable DETR \cite{deformable}, DN-DETR \cite{dn}, DINO\cite{dino}, Co-DETR \cite{codetr}, RT-DETR \cite{rtdetr}. Despite its outstanding results in several detection benchmarks, DETR has 2 main problems: its poor convergence and the "black box" meaning of queries. Many methods have been proposed, such as alternating attention mechanism \cite{deformable}, introducing new query structure (positional query and content query) 
 \cite{dab}, reducing instability in Hungarian matching by: denoising queries \cite{dn}, constrastive queries \cite{dino}, one-to-many labels positive queries \cite{codetr}.

Besides, in continual learning, there are 2 most common strategies to reduce catastrophic forgetting, which are Knowledge Distillation (KD): ~\cite{LiH2018LwF,Douillard2020PODNet,Zhao2020Maintaining} and Exemplar Replay (ER): ~\cite{Rebuffi2017iCaRL,Liu2020Mnemonics,Wang2022Memory,Castro18EndToEnd}. To be specific, Knowledge Distillation applies regularization to maintain prior knowledge while training the model in new tasks, this helps preserve feature maps of the model in new tasks resembling its upstream tasks. In addition, Exemplar Replay is a way to store old data from previous tasks in buffers, that models can use in current tasks to prevent forgetting.

In comparison with traditional continual learning, incremental object detection requires models not only to classify objects into their corresponding classes but also to detect their positions in images (their bounding boxes). Furthermore, it also differs from standard detection, in that models need to be able to train incrementally while avoiding catastrophic forgetting upstream tasks’ knowledge. 

Due to its importance and difficulties, incremental object detection has attracted considerable research interest.  CL-DETR \cite{cldetr} has shown that directly applying ER and KD methods of continual learning to detection models is not effective. Therefore, they proposed new methods of KD in object detection by pseudo labeling: using a model in the first task (task 0) to generate the most confident foreground predictions as pseudo labels and merge them with the ground-truth labels of the current task. Besides, there exists a notable method of ER in continual object detection such as: SDDGR \cite{sddgr} used diffusion models as a new way of data-replaying. Such methods suffer from both poor performance and low computational efficiency in inference time, which is impractical in real-life implementations. In addition, despite the semantic relationships between classes often forming a hierarchical structure, recent studies have not effectively utilized this inherent characteristic, particularly in the context of object detection. In contrast, assuming that we already have the hierarchical information of all classes’ labels, leveraging it to enhance performance while maintaining efficiency in the continual learning settings is more challenging.

To address the above problems, we propose Hier-DETR, a model for incremental object detection that achieves a good balance between high performance and efficiency during inference. Firstly, we use a new backbone: RT DETR \cite{rtdetr}, and adapt it to CL settings. Though, to our best knowledge, Co-DERT \cite{codetr} is the SOTA of the object detection model but is not efficient for real-time running, RT-DETR is still the first real-time end-to-end object detector, that can maintain equilibrium between speed and accuracy.  Secondly, we introduce a novel technique namely Hierarchical Neural Collapse (HNC) for boosting both the performance and efficiency of OD. Using neural collapse (NC) to improve efficiency \cite{ncfscil,yang2022inducingneuralcollapseimbalanced} is an emerged technique, that pre-assigns fixed prototypes as the structure of simplex Equiangular Tight Frame  (ETFs)  for each class and preserves them as consistent prior throughout incremental learning.  Nevertheless, the implementation of neural collapse within a hierarchical structure has not been investigated. We propose a strategy to initialize Hierarchical Neural Collapse in incremental object detection. Our approach is proven to alleviate catastrophic forgetting by retaining information from upstream tasks while simultaneously balancing with downstream knowledge thanks to the characteristics of NC. Additionally, our experiments also indicate that integrating HNC into various DETR-based models reduces the instability of queries during the Hungarian matching process, thereby contributing to faster convergence.

We empirically demonstrate the advantage of Hier-DETR in MTSD (Mapillary Traffic Sign Dataset) \cite{ertler2020mapillary} and COCO \cite{coco} datasets with various settings. The results show that Hier-DETR outperforms \textbf{48.4} in COCO and \textbf{45.46} in MSTD datasets in comparison with recent SOTAs. The ablation study also indicates that Hierarchical NC boosts models’ stability in Hungarian Matching.

To summarize, our main contributions are as follows:
\begin{enumerate}
    \item We proposed Hier-DETR, an innovative continual learning strategy for object detection based on NC and a hierarchical structure of classes. The method can be easily applied to any DETR-based model to achieve outstanding performance and faster convergence.
    \item To the best of our knowledge, this paper is the first to apply the hierarchical semantic meaning of classes' labels with NC. We expect that this provides a different perspective for solving the problem of forgetting in continual learning, especially in COD.
    \item We conducted several experiments to demonstrate the effectiveness of HNC in various settings on multiple COD datasets, especially on MTSD, a traffic signs dataset where all labels have strict relationships. Our method outperforms all previous SOTA incremental detection models on all benchmarks.
\end{enumerate}

\section{Related Work}
\subsection{Transformer-based Object Detectors}

Thanks to the remarkable performance of Transformer \cite{attention} in natural language processing, researchers have attracted attention to explore the implementation of Transformer architecture for vision tasks. Carion et al.\cite{detr} first proposed end-to-end object detector based on Transformers (DETR) which eliminates the need for hand-crafted anchor and NMS components. Although DETR has achieved competitive performance, it suffers from the slow convergence problem, computationally intensive, and the role of queries in Decoder. Recently, many works have focused on accelerating the training process of DETR. Deformable-DETR \cite{deformable} designed a new attention mechanism that only attends to a fixed number of points (reference points). Conditional DETR \cite{conditional} and DAB-DETR \cite{dab} separated each decoder query into a content part and a position part as an anchor box to utilize each part effectively. 

Another line of work considers severe challenges arising from the instability of Hungarian Matching algorithms. DN-DETR \cite{dn} introduces a novel training method to accelerate DETR training by denoising with attention mask. They divides queries into 2 components: denoising part (comprising multiple denoising groups created from different noise versions added to each ground-truth object) and the matching part. Following this, DINO \cite{dino} proposed Contrastive DeNoising Training to enhance the model's capability of inhibiting confusion of multiple anchors referring to the same object by "negative queries" (denoising queries assigned to "background"). Currently, the state-of-the-art of DETR-based model: CO-DETR \cite{codetr} proposes Customized Positive queries, which are generated through one-to-one label assignment from the encoder's output. 

Besides, all previous works are still heavily computational and overlook the challenge of real-time inference. RT-DETR \cite{rtdetr} is the only Transformer-based detection model, that can detect in real-time while still ensuring competitive performance.

\subsection{Class incremental learning}

Class-incremental learning (CIL) is a branch of continual learning, that incorporates models to integrate new classes in training phases and to preserve performance on upstream tasks. Class incremental learning approaches can be organized into 2 main categories which are  Knowledge Distillation (KD) ~\cite{LiH2018LwF,Douillard2020PODNet,Zhao2020Maintaining} and Exemplar Replay (ER) ~\cite{Rebuffi2017iCaRL,Liu2020Mnemonics,Wang2022Memory,Castro18EndToEnd}. Knowledge Distillation (KD) aims to retain the knowledge captured by the models in previous tasks by aligning its logits or its feature maps. Exemplar Replay (ER) is a rehearsal with stored data as "buffer" or sampling from a generative model in order to replay data, and knowledge of previous tasks, that can mitigate forgetting.

\subsection{Class incremental object detection}

Class incremental object detection is more challenging than standard class incremental classification, as an image can contain multiple ground-truth objects. During each continual phase, both old and new objects may appear within the same image. Existing approaches have investigated both Knowledge Distillation (KD) and Exemplar Replay (ER) techniques for conventional detectors (Faster-RCNN~\cite{faster} and GFL~\cite{gfl}) under continual learning settings. Shmelkov et al.~\cite{Shmelkov2017Incremental} first introduce KD to the output of Faster-RCNN. Following this line, other methods apply KD to intermediate feature maps~\cite{Yang2022Multi,Zhou2020LifelongOD,Feng2022ElasticResponse} and region proposal networks~\cite{Chen2019New,Hao2019EndtoEnd,Peng2020FasterILOD}. Other approaches such as \cite{Joseph2021TowardsOpenWorld, Dai2021UPDETR} proposed novel sampling techniques for selecting higher-quality exemplars to facilitate model fine-tuning after each training phase. 

Futhermore, most recent state-of-the-art models in class incremental object detection: CL-DETR~\cite{cldetr} and SDDGR~\cite{sddgr} introduced strategies for integrating KD and ER to DETR-based models. In particular, CL-DETR~\cite{cldetr} identifies the limitations of directly applying KD and ER to detection transformer models, then they proposed novel KD techniques: pseudo-labeling. SDDGR~\cite{sddgr} introduced the use of diffusion model as an innovative approach to data replay.

\subsection{Neural Collapse}

Papyan et al. \cite{neural_collapse} reveals the neural collapse phenomenon, that the last-layer features converge to their within-class means, and the within-class means together with the classifier vectors collapse to the vertices of a simplex equiangular tight frame at the terminal phase of training on a balanced dataset.

\begin{definition}[Simplex Equiangular Tight Frame]
	\label{ETF}
	A collection of vectors $\mathbf{m}_i \in \mathbb{R}^d$, $i=1,2,\cdots,K$, $d\ge K-1$, is said to be a simplex equiangular tight frame if:
	\begin{equation}\label{ETF_M}
		\mathbf{M} = \sqrt{\frac{K}{K-1}} \mathbf{U} \left( \mathbf{I}_K - \frac{1}{K} \mathbf{1}_K \mathbf{1}_K^T \right),
	\end{equation}
	where $\mathbf{M} = [\mathbf{m}_1, \cdots, \mathbf{m}_K] \in \mathbb{R}^{d\times K}$, $\mathbf{U} \in \mathbb{R}^{d\times K}$ allows a rotation and satisfies $\mathbf{U}^T \mathbf{U} = \mathbf{I}_K$, $\mathbf{I}_K$ is the identity matrix, and $\mathbf{1}_K$ is an all-ones vector. 
\end{definition}

All vectors in a simplex ETF have an equal $\ell_2$ norm and the same pair-wise angle, \emph{i.e.,}
\begin{equation}\label{mimj}
	\mathbf{m}_i^T \mathbf{m}_j = \frac{K}{K-1}\delta_{i,j} - \frac{1}{K-1}, \forall i, j\in[1,K],
\end{equation}
where $\delta_{i,j}$ equals 1 when $i=j$ and 0 otherwise. The pair-wise angle $-\frac{1}{K-1}$ is the maximal equiangular separation of $K$ vectors in $\mathbb{R}^d$ \cite{neural_collapse}. 

Then the neural collapse (NC) phenomenon can be formally described as:

\vspace{0.2cm}

\textbf{(NC1)} Within-class variability of the last-layer features collapse: $\Sigma_W\rightarrow\mathbf{0}$, and $\Sigma_W := \mathrm{Avg}_{i,k}\{(\h_{k,i}-\h_k)(\h_{k,i}-\h_k)^T\}$, where $\h_{k,i}$ is the last-layer feature of the $i$-th sample in the $k$-th class, and $\h_k = \mathrm{Avg}_{i}\{\h_{k,i}\}$ is the within-class mean of the last-layer features in the $k$-th class;

\vspace{0.2cm}

\textbf{(NC2)} Convergence to a simplex ETF: $\tilde{\h}_k = (\h_k-\h_G)/||\h_k-\h_G||, k\in[1,K]$, satisfies Eq. (\ref{mimj}), where  $\h_G$ is the global mean of the last-layer features, \emph{i.e.,} $\h_G = \mathrm{Avg}_{i,k}\{\h_{k,i}\}$;

\vspace{0.2cm}

\textbf{(NC3)} Self duality: $\tilde{\h}_k=\w_k/||\w_k||$, where $\w_k$ is the classifier vector of the $k$-th class;

\vspace{0.2cm}

\textbf{(NC4)} Simplification to the nearest class center prediction: $\argmax_k\langle\h, \w_k\rangle=\argmin_k||\h-\h_k||$, where $\h$ is the last-layer feature of a sample to predict for classification.

\begin{definition}[General Orthogonal Frame]
\label{def:GOF}
    A standard general orthogonal frame (GOF) is a collection of points in $\mathbb{R}^{K}$ specified by the columns of:
    \begin{align}
        \mathbf{N} = \frac{1}{\sqrt{\sum_{k=1}^{K} a_{k}^{2}}} \operatorname{diag}(a_{1}, a_{2}, \ldots, a_{K}), \: a_{i} > 0 \: \: \forall \: i \in [K]. \nonumber
    \end{align}

    Where $a_{i}$ denotes the number of data points associated with the corresponding class in the dataset. They also examined the general version of GOF~\cite{hienneuralcollapse} as a collection of points in $
    \mathbb{R}^{d} \: (d \geq K)$ specified by the columns of $ \mathbf{P} \mathbf{N}$ where $\mathbf{P} \in \mathbb{R}^{d \times K}$ is an orthonormal matrix, i.e. $\mathbf{P}^{\top} \mathbf{P} = \mathbf{I}_{K}$.
    
\end{definition}
However, NC only occurs when the dataset is balanced, which is not suitable for most object detection datasets. Hien et al. \cite{hienneuralcollapse} proved that the convergence of the last-layer features and classifiers to a geometry, called GOF (\ref{def:GOF}), consisting of orthogonal vectors, whose lengths depend on the amount of data in their corresponding classes.

%
%
%
%
%

\section{Proposed method}

\subsection{Overview}
Following the standard incremental object detection setting \cite{cldetr}, the model is continually shown N tasks w.r.t. mutually exclusive subsets of object classes' categories. To clarify, let $D = {(x, y)}$ represent a dataset consisting of images $x$ and their corresponding object labels $y$ and let $C = \{1, \dots, C\}$ denote the complete set of object categories. Then, Liu et al. \cite{cldetr} divided the sets $D$ and $C$ into $M$ subsets: $D = D_1 \cup \dots \cup D_M$ and $C = C_1 \cup \dots \cup C_M$, corresponding to each training phase. For each phase $i$, they adjusted the samples $(x, y) \in D_i$ so that $y$ only retains the annotations for objects of class $C_i$. Note that images can contain objects that belong to any type of $C$, but only types of $C_i$ are annotated during every training phase $i$. This strict setting has been proven to maintain all characteristics of normal incremental learning in object detection.

We introduce an innovative efficiency framework for IOD problems, named Hier-DETR, via Hierarchical Neural Collapse construction (Section~\ref{hnc_construction}), Guiding query to HNC structure by Proxy-NCA \cite{proxynca} (Section~\ref{proxy_loss}).

\subsection{Hierarchical Neural Collapse Construction}
\label{hnc_construction}
 \begin{figure*}[!t]
	\begin{center}
		\includegraphics[width=0.9\linewidth]{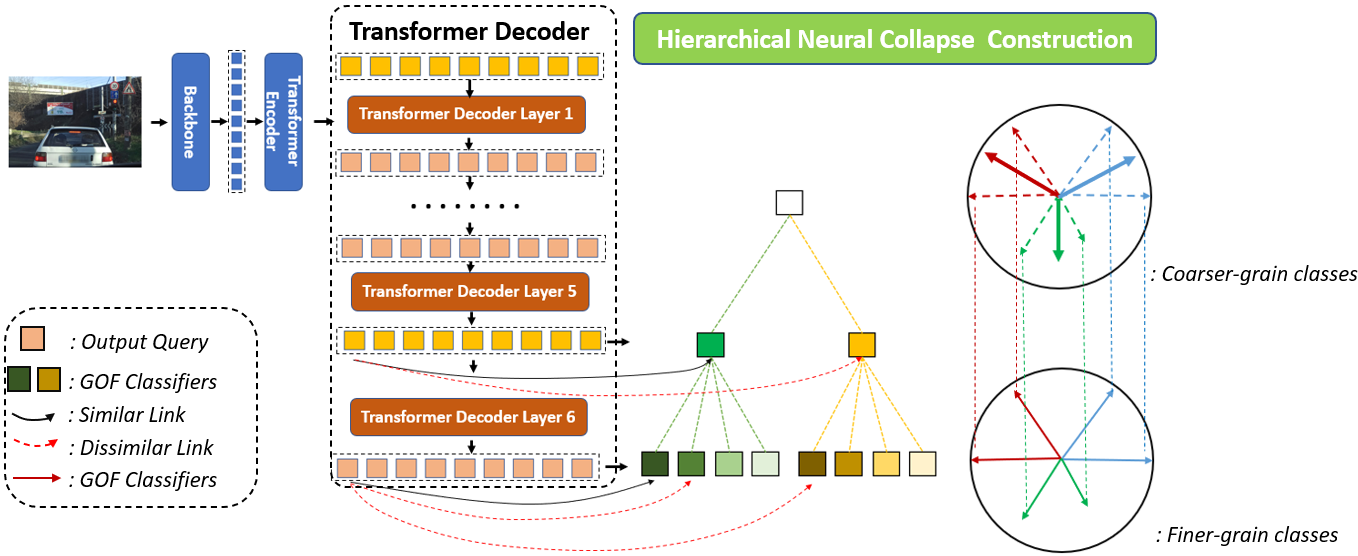}
		\vspace{-2.mm}
		\caption{\textbf{Overview of Hier-DETR:} Firstly, the image is processing through backbone (ResNet50) and RT-DETR Encoder to produce feature for decoder layers. Then, we applies HNC to guide each query to its corresponding coarser-grain and finer-grain classes. Due to characters of proxy loss, after labeled by Hungarian Matching, each query is "pulled" close to others if they have same (super) class in (early) decoder layer, "pushed" away to different-class queries.}
		\label{model_structure}
	\end{center}
	\vspace{-3mm}
\end{figure*}

\label{Hierachical Neural Collapse Construction}
While using neural collapse as fix classes' prototypes has shown exceptional performance in classification tasks, particularly in fine-grain and few-shot scenarios, existing methods do not utilize the semantic meaning of labels, as this can easily disrupt the characteristics of Neural Collapse (NC). Intuitively, we desire a method that allows the meaning of class labels to influence the neural collapse structure without conflicting with any NC attributes and maintains training efficiency.

Specifically, we replace conventional Neural Collapse (NC) fixed class prototypes \cite{yang2022inducingneuralcollapseimbalanced,ncfscil} with a hierarchical \textbf{"tree"} structure, which consists of numerous layers of General Orthogonal Frames (GOFs~\ref{def:GOF}) corresponding to different levels of semantic meaning in class labels.

\textbf{In each training phase $j$}, with the assumption that hierarchical information exists for all class labels, we denote classes' tree as $T$, which have $m$ levels $L_i$, each corresponding to a specific semantic level of all classes in $K_j$, where $K_j= C_1 \cup \dots \cup C_j$.

Specifically, let $\mathbf{\W_{GOF_1}}$ represent the General Orthogonal Frame (GOF~\ref{def:GOF}) prototypes of the first layer in the Hierarchical Neural Collapse structure (HNC), defined as $\mathbf{\W_{GOF_1}} = [\mathbf{m}_{11}, \cdots, \mathbf{m}_{1K}] \in \mathbb{R}^{d \times K}$, where $\mathbf{m}_{1i}\in \mathbb{R}^{d}$ denotes the column vectors of $\mathbf{\W_{GOF_1}}$ for all $1 \le i \le |L_1|$ and $K = |L_1|$, where $|L_1|$ is the number of class in $L_1$. During task $j$, a new set of vectors $[\mathbf{m}_{1{(K+1)}}, \cdots, \mathbf{m}_{1{(K+|C_j|)}}]$ corresponding to each class in the new class set $C_j$ can be initialized while satisfying the following conditions: each vector is orthogonal to all others, as well as to every vector in $\mathbf{\W_{GOF_1}}$, and its $L_2$ norm is proportional to the number of samples of that class in the dataset.

\begin{figure*}[!t]
	\begin{center}
		\includegraphics[width=1\linewidth]{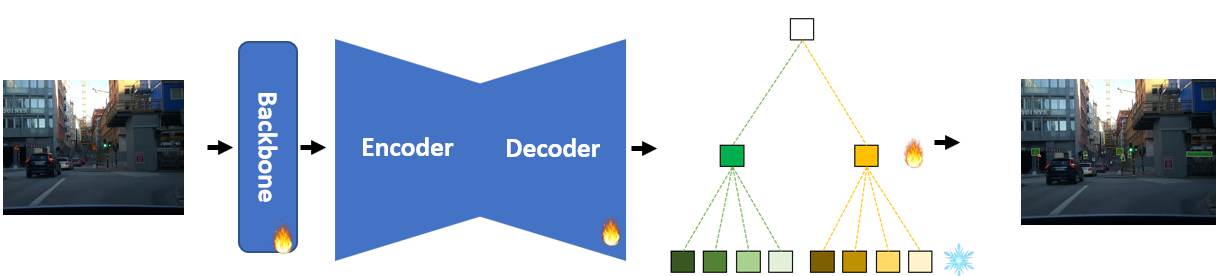}
		\vspace{-2.mm}
		\caption{\textbf{An illustration of Hier-DETR in CL pipeline}. During CL phases, only each assigned class prototype in the first layer (L1) of HNC fixed as knowledge distillation. When new classes emerge, they can be added to the first layer. All other subsequent layers of HNC structure adapt based on the expansion of the whole class's category set.}
            \label{cl pipeline}
	\end{center}
	\vspace{-3mm}
\end{figure*}

Then, each subsequent layer $i^{th}$ is initialized based on the structure of $\W_{GOF_{i-1}}$ with following conditions:
\begin{enumerate}
    \item \( W_{GOF_{i}} \) is a GOF structure with \( K = |L_i| \).
    
    \item Each column vector \( W_i \) of \( W_{GOF_{i}} \) represents a super-class in level $L_i$ of the class hierarchy. Its value equals to the mean of every column vector of child-class of \( W_{GOF_{i-1}} \) in level $L_{i-1}$. Then, each of these vectors is normalized and scaled up to their corresponding number of samples in its "class". 
    
    \item Every assigned class prototype in the first layers ($L_1$) of HNC is fixed during training. Whenever novel classes appear, they can be added to HNC's first layer without breaking the characteristics of GOFs (according to Theorem~\ref{theorem}). For all continual learning phases, the following layers are adjusted depending on the growth of the whole set of classes. (Fig.\ref{cl pipeline})
    
    \item As the tree gradually grows, HNC adapts each GOF layer structure while still preserving its characteristics throughout training phases.
\end{enumerate}

After the output of the decoder's query is labeled by the Hungarian Matching algorithm \cite{Kuhn1955Hungarian}, these outputs are considered as the "last-layer features" of objects. Subsequently, they are aligned with the Hierarchical Neural Collapse (HNC) prototypes corresponding to their assigned labels. We denote output of querry $q$ through $l$ decoder layers as $output_q={q_1, q_2,\dots, q_l}$, with its assigned class as $c_q$,  where the class hierarchy is represented as ${c_{q_1},\dots,root }$. Each feature in $output_q$ is matched to every vector in HNC corresponding to each label in its class family.

\textbf{Comparison between existing methods using normal neural collapse \cite{yang2022inducingneuralcollapseimbalanced,ncfscil} and our HNC structure}: 

Although normal neural collapse prototypes have been proven to be "maximally pairwise-distanced" from each other, Papyan et al. \cite{neural_collapse} demonstrated that this property only holds when the dataset is perfectly balanced. However, most datasets, especially in object detection, are extremely imbalanced, where the number of samples for some object classes far exceeds others. Nevertheless, Hien et al. \cite{hienneuralcollapse} proved that when data is imbalanced, the class-means matrix and the last-layer classifiers still converge to the General Orthogonal Frames (GOFs). Moreover, existing methods \cite{yang2022inducingneuralcollapseimbalanced, ncfscil} based on normal neural collapse as fixed classifiers cannot handle with the challenge of expanding number of classes in continual learning settings. Specifically, models must be initialized with a fixed $K$ classes to satisfy condition (\ref{ETF_M}), which contradicts to the purpose of continual learning settings. In terms of performance, Yang et al.~\cite{ncfscil} also claimed that when using the OF (orthonormal frame) structure, the model's performance just slightly changed. Consequently, the GOFs structure is the most optimal fixed architecture for class classifiers in incremental object detection, both theoretically and experimentally.

\begin{theorem}
	\label{theorem}
Let $\mathbf{W_{GOF_i}}$ denote the General Orthogonal Frame (GOF~\ref{def:GOF}) prototypes of the $i^{\text{th}}$ layer in the Hierarchical Neural Collapse (HNC) structure. We have $\mathbf{W_{GOF_i}} = [\mathbf{m}_{i1}, \cdots, \mathbf{m}_{i|L_i|}] \in \mathbb{R}^{d \times |L_i|}$, where $\mathbf{m}_{it} \in \mathbb{R}^{d}$ represents the column vectors of $\mathbf{W_{GOF_i}}$ for all $1 \le t \le |L_i|$, with $|L_i|$ as the number of classes in $L_i$. Suppose that $\mathbf{W_{GOF_{i-1}}} = [\mathbf{m}_{(i-1)1}, \cdots, \mathbf{m}_{(i-1)|L_{i-1}|}] \in \mathbb{R}^{d \times |L_{i-1}|}$ represents the GOFs of the $(i-1)^{\text{th}}$ layer in the HNC structure, and we have the class taxonomy with respect to the $i^{\text{th}}$ layer:
\begin{align}
\left\{ \mathbf{m}_{it}: \left[\mathbf{m}_{(i-1)k} \;|\; \mathbf{m}_{(i-1)k} 
\in \mathbf{Child}_{\mathbf{m}_{it}} \right] \right\}, \quad 
\end{align}
for all \( 1 \le t \le |L_i| \), where \( \mathbf{Child}_{\mathbf{m}_{it}} \) denotes the set of all child classes of \( \mathbf{m}_{it} \) in the $(i-1)^{\text{th}}$ layer.

Then, each vector $\mathbf{m}_{it}$ is orthogonal to each other.

\textbf{Proof:} Consider two vectors $\mathbf{m}_{it_1}$ and $\mathbf{m}_{it_2}$. We have:

\[
\cos \angle(\mathbf{m}_{it_1}, \mathbf{m}_{it_2}) 
= \mathbf{m}_{it_1}^T \mathbf{m}_{it_2} 
\]



{\fontsize{6}{6} \selectfont
\[
= \frac{\sum_{\mathbf{m}_{(i-1)k} 
\in\mathbf{Child}_{\mathbf{m}_{it_1}}} \mathbf{m}_{(i-1)k}}{|\mathbf{Child}_{\mathbf{m}_{(i-1)t_1}}|} 
\cdot \frac{\sum_{\mathbf{m}_{(i-1)k} 
\in\mathbf{Child}_{\mathbf{m}_{it_2}}} \mathbf{m}_{(i-1)k}}{|\mathbf{Child}_{\mathbf{m}_{(i-1)t_2}}|} 
\]
}
=0,

Given that $\mathbf{m}_{(i-1)k_1}^T \mathbf{m}_{(i-1)k_2} = 0$ for all $k_1, k_2 \in [1, \cdots,|L_{i-1}|]$ (due to characteristic of GOFs in the $(i-1)^{th}$ layer).

\end{theorem}

\subsection{Guiding query to HNC structure by Proxy-NCA}

\label{proxy_loss}

Instead of only using neural collapse to the last decoder layer\cite{yang2022inducingneuralcollapseimbalanced,ncfscil}, we allocated each query to its HNC prototypes across the outputs of several decoder layers.

In particular, in the $i^{th}$ decoder layer, 
\begin{align}
    L(q_i) & = \sum_{q_i \in Q_i} -\log \frac{e^{d(q_i, w_j^+)}}
        {\displaystyle\sum\limits_{w_j^{-} \in W_{GOF_i^{-}}} e^{d(q_i, w_j^-)}} \\
    & = \sum_{q_i \in Q_i} \Bigg\{ -d(q_i, w_j^+) + \log \left( \sum_{w_j^{-} \in W_{GOF_i^{-}}} e^{d(q_i, w_j^-)} \right) \Bigg\} \\
    & = \sum_{q_i \in Q_i} \Bigg\{ -d(q_i, w_j^+) + \underset{w_j^{-} \in W_{GOF_i^{-}}}{\mathrm{LSE}} \, d(q_i, w_j^-) \Bigg\}, \label{eq:Proxy_NCA_reform}
\end{align}
Where $w_j^+$ and $w_j^{-}$ denote positive and negative proxies, respectively; $q_i$ represents the specific output of an individual query; and $Q_i$ denotes the set of all query outputs produced by the $i^{th}$ decoder layer. $W_{GOF_i^{-}}$ is set of all negative proxies (all different classes from the class of $q_i$) and $W_{GOF_i^{-}}$ is the set of positive proxies (the same class as $q_i$) in $W_{GOF_i}$.

The gradient of Proxy-NCA loss with respect to $d(q_i,w)$ is given by

\begin{align}
    \frac{\partial L(q_i)}{\partial d(q_i,w)} =
    \begin{cases}
    -1, & \text{If } w = w_j^+, \\
    \frac{e^{d(q_i,w)}}{\sum\limits_{w_j^{-} \in W_{GOF_i^{-}}} e^{d(q_i,w_j^-)}}, & \text{Otherwise.}
    \end{cases}
    \label{eq:Proxy_NCA_grad}
\end{align}

By aligning each output of the same query across different decoder layers, we can preserve not only the structure of GOF in each layer in HNC but also the hierarchical semantic relationships among class labels. Thanks to HNC, each query can still learn features from all classes at the initial decoder layers and gradually focus on the attribute of its "final" predicted class. Specifically, due to the direction of the gradient of Proxy-NCA loss, at the very first decoder layer, all representations of queries assigned to the same "super-class" are pulled to be closed to each other and to their GOF prototypes, while being pushed apart from prototypes of other super-classes, then. Until the final decoder layer, HNC guides each query toward its corresponding finer-coarse class. In contrast, directly applying ETF prototypes to the last decoder layer merely "forces" the queries' feature to to align with arbitrary ETF prototypes. This can lead to critical convergence issues, especially when taking gradients, as the randomly initialized prototypes cannot be properly adjusted to match the query features' lengths.

\section{Experiment}

\begin{table*}[htbp]
  \centering
  \resizebox{\textwidth}{!}{%
    \begin{tabular}{lcccccc|cccccc}
      \toprule
      \textbf{Method} & \multicolumn{6}{c|}{\textbf{70 + 10 Setting}} & \multicolumn{6}{c}{\textbf{40 + 40 Setting}} \\
      \cmidrule(lr){2-7} \cmidrule(lr){8-13}
       & AP & AP$_{50}$ & AP$_{75}$ & AP$_S$ & AP$_M$ & AP$_L$ 
       & AP & AP$_{50}$ & AP$_{75}$ & AP$_S$ & AP$_M$ & AP$_L$ \\ 
      \midrule
      LWF     & 7.1  & 12.4 & 7.0  & 4.8  & 9.5  & 10.0  & 17.2 & 45.0 & 18.6 & 7.9  & 18.4 & 24.3 \\
      RILOD   & 24.5 & 37.9 & 25.7 & 14.2 & 27.4 & 33.5  & 29.9 & 45.0 & 32.0 & 15.8 & 33.0 & 40.5 \\
      SID     & 32.8 & 49.0 & 35.0 & 17.1 & 36.9 & 44.5  & 34.0 & 51.4 & 36.3 & 18.4 & 38.4 & 44.9 \\
      ERD     & 34.9 & 51.9 & 37.4 & 18.7 & 38.8 & 45.5  & 36.9 & 54.5 & 39.6 & 21.3 & 40.4 & 47.5 \\
      CL-DETR & 40.4 & 58.0 & 43.9 & 23.8 & 43.6 & 53.5  & 42.0 & 60.1 & 45.9 & 24.0 & 45.3 & 55.6 \\
      SDDGR   & 40.9 & 59.5 & 44.8 & 23.9 & 44.7 & 54.0  & 43.0 & 62.1 & 47.1 & 24.9 & 46.9 & 57.0 \\
      \textbf{Ours} 
              & \textbf{48.4} & \textbf{65.9} & \textbf{52.9} & \textbf{31.2} & \textbf{51.3} & \textbf{70.3}  
              & \textbf{47.1} & \textbf{63.6} & \textbf{51.2} & \textbf{30.0} & \textbf{50.8} & \textbf{61.7} \\
      \bottomrule
      \label{mtsd}
    \end{tabular}
  }
  \caption{CIOD results (\%) on COCO 2017 in two-phase settings. All models' performance of LWF \cite{LiH2018LwF}, RILOD \cite{Li2019RILOD}, SID \cite{Peng2021SID}, ERD \cite{feng2022overcoming}, CL-DETR \cite{cldetr}, SDDGR \cite{sddgr} are extracted from SDDGR \cite{sddgr}. The best performance is emphasized in \textbf{bold}, and a red upward arrow ( {\color{red}↑}) denotes an enhancement relative to the prior state-of-the-art approaches.}
  \label{coco_comparison}
\end{table*}

\begin{table*}[htbp]
  \centering
  \resizebox{\textwidth}{!}{%
    \begin{tabular}{lcccccc|cccccc}
      \toprule
      \textbf{Method} & \multicolumn{6}{c|}{\textbf{129 + 92 Setting}} & \multicolumn{6}{c}{\textbf{150 + 71 Setting}} \\
      \cmidrule(lr){2-7} \cmidrule(lr){8-13}
       & AP & AP$_{50}$ & AP$_{75}$ & AP$_S$ & AP$_M$ & AP$_L$ 
       & AP & AP$_{50}$ & AP$_{75}$ & AP$_S$ & AP$_M$ & AP$_L$ \\ 
      \midrule
      CL-DETR & 33.8 & 49.6 & 36.6 & 18.1 & 38.4 & 54.8  
             & 36.5 & 52.3 & 39.7 & 19.6 & 40.1 & 58.2 \\
      SDDGR   & 34.6 & 51.1 & 38.4 & 18.8 & 39.7 & 56.9  
             & 37.4 & 56.4 & 40.6 & 21.8 & 38.7 & 59.1 \\
      CL-RTDETR & 41.7 & 57.6 & 47.2 & 19.3 & 39.8 & 68.7  
                & 43.13 & 59.0 & 48.5 & 20.7 & 42.6 & 71.8  \\
      \midrule
      \textbf{Ours} 
              & \textbf{43.2} & \textbf{58.9} & \textbf{48.7} & \textbf{20.1} & \textbf{42.7} & \textbf{72.4}  
              &\textbf{45.46} & \textbf{59.1} & \textbf{50.8} & \textbf{19.4} & \textbf{44.3} & \textbf{74.3}  \\
      \bottomrule
    \end{tabular}
  }
  \caption{CIOD results (\%) on MTSD in two-phase settings. We reproduce all previous state-of-the-art models: CL-DETR \cite{cldetr} and SDDGR \cite{sddgr} on MTSD - traffic-sign dataset. We also implemented RT-DETR as a new baseline for CL-DETR (instead of Deformable-DETR \cite{deformable} to show the benefit of the HNC structure. The best performance is emphasized in \textbf{bold}, and a red upward arrow ( {\color{red}↑}) denotes an enhancement relative to the prior state-of-the-art approaches.}
  \label{mtsd_per}
\end{table*}

\subsection{Dataset and metrics}
We benchmark our work with different settings on common datasets: COCO \cite{coco} and MTSD (Mapillary Traffic Sign Dataset) \cite{mtsd}.\textbf{MS COCO 2017} \cite{coco} is a widely used dataset that consists of 115K labeled images for training and 5K images for validation. \textbf{MTSD} (Mapillary Traffic Sign Dataset) is a diverse street-level imagery dataset featuring bounding box annotations for the detection and classification of 400 different classes of traffic signs worldwide. Especially, MTSD is an object detection dataset, with class labels that adhere to a strict hierarchical structure~\cite{hierarchicalmtsd}. Both of these datasets are extremely imbalanced.

For evaluation, we report the standard average precision (AP) results on COCO \texttt{val2017} and MTSD \texttt{val} under different intersection over union (IoU) thresholds (\textit{AP}, \textit{AP.5}, \textit{AP.75}) and object scales (\textit{AP}$_S$, \textit{AP}$_M$, \textit{AP}$_L$). In the ablation study, for evaluating instability in Hungarian Matching, we compare the instability \textit{IS} \cite{dab} of the original RT-DETR and our Hier-DETR. 

\subsection{ Implementation and experiments}

\quad \textbf{Implementation Details}: 

Our work is built upon RT-DETR which uses ImageNet pre-trained ResNet-50 for backbone, 300 queries, and 100 denoising queries. For COCO 2017, we keep all the standard settings with the baseline \cite{rtdetr}. For MTSD, we reproduce CLDETR \cite{cldetr} and SDDGR \cite{sddgr} to compare with our approach under 2 continual learning schemas. Furthermore, we implement RT-DETR~\cite{rtdetr} as a baseline model (instead of Deformable-DETR~\cite{deformable}) within  CL-DETR~\cite{cldetr} framework, called CL-RTDETR. This modification allows us to investigate the impact of HNC on transformer-based detection models in incremental object detection. All experiments are performed using NVIDIA GeForce RTX 3080 GPUs with a batch size of 8. To maintain the equality between previous models, we train all the models for 50 epochs in every training phase (task).

\vspace{0.2cm}

\textbf{Experiment setups}:

We conduct our framework on several scenarios of the \textit{two-phase setting}. In detail, we train our model for 2 tasks: Task 0 for $T_0$ classes, Task 1 for $T_1$ classes, and evaluating in total $T_0+T_1$ classes. For instance, there are 2 settings for COCO ("70+10" and "40+40") and 2 settings for MTSD ("150+71" and "129+92"). 

\textbf{Experiments on COCO dataset:}

Table~\ref{coco_comparison} demonstrates that our proposed method achieves remarkable performance across all AP metrics, surpassing previous approaches by substantial margins such as: 7.5\% in $AP$ for "70+10" and 4.1\% in $AP$ for "40+40" compared to SDDGR \cite{sddgr}. Although the class hierarchy of the COCO dataset is a relatively simple '3-level tree' structure, with 80 object categories grouped in 12 superclasses, our method's Hierarchical Neural Collapse (HNC) approach still demonstrates a notable improvement in performance, even in cases where the semantic relationships between labels do not strictly adhere to a hierarchical structure.
\vspace{0.2cm}

\textbf{Experiments on MTSD dataset:}

\vspace{0.1cm}

Table~\ref{mtsd_per} shows that our method outperforms across all AP metrics especially: 8.6\% in $AP$ for "129+92" and 8.06\% in $AP$ for "150+71" compared to previous state-of-the-art (SDDGR \cite{sddgr}). The experiments show that with the help of the HNC structure, each query has been guided gradually from its coarser-class (super-classes) to its finer-class (final predicted class). 

To some extent, we use a different baseline compared to CL-DETR \cite{cldetr}, and SDDGR \cite{sddgr}. However, we have reproduced CL-DETR~\cite{cldetr} with RT-DETR~\cite{rtdetr} as its baseline model for a more fair comparison. Furthermore, our Hier-DETR continues to demonstrate a 1.5\% improvement in $AP$ for ``129+92'' and a 4.1\% improvement in $AP$ for ``150+71'' when compared to CL-RTDETR (CL-DETR utilizing RT-DETR as the baseline). Consequently, we can check the adaptability of our strategy in several scenarios.

We also implement our Hier-DETR in join-training (normal training setting) for MTSD dataset. In particular, we maintain all the default settings from the RT-DETR~\cite{rtdetr} configuration and train both RT-DETR and our Hier-DETR for 50 epochs in MTSD dataset. Table~\ref{join_training} illustrates that our approach achieves superior performance compared to the currently competitive model.

\begin{table}[htbp]
  \centering
  {%
    \begin{tabular}{lcccccc}
      \toprule
      \textbf{Method} & \multicolumn{6}{c}{\textbf{Join-training Setting}}  \\
      \cmidrule(lr){2-7} 
       & AP & AP$_{50}$ & AP$_{75}$ & AP$_S$ & AP$_M$ & AP$_L$  \\ 
      \midrule
      RT-DETR & 38.4 & 51.3 & 43.6 & 14.4 & 38.8 & 68.7  \\
      
      \midrule
      \textbf{Ours} 
              & \textbf{39.8} & \textbf{52.9} & \textbf{45.6} & \textbf{15.5} & \textbf{40.5} & \textbf{70.5}    \\
      \bottomrule
    \end{tabular}
  }
  \caption{Join-training results of RT-DETR~\cite{rtdetr} and our Hier-RTDETR on MTSD dataset.}
  \label{join_training}
\end{table}

\begin{figure*}[!t]
    \centering
    \begin{subfigure}[t]{0.4\textwidth}
        \centering
        \includegraphics[width=\textwidth]{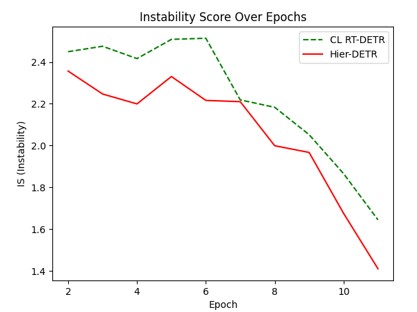}
        \caption{}
        \label{instability1}
    \end{subfigure}
    \hfill
    \begin{subfigure}[t]{0.55\textwidth}
        \centering
        \includegraphics[width=\textwidth]{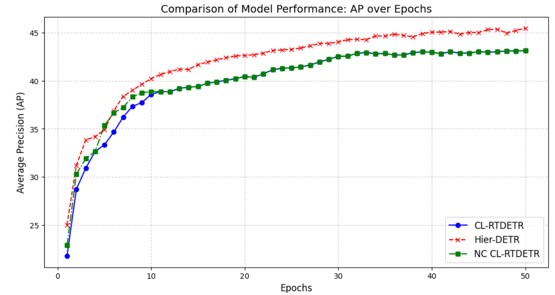}
        \caption{}
        \label{instability2}
    \end{subfigure}
    \caption{a): The IS of Hier-DETR and CL-RTDETR during 12 epochs of training on task 1 of the "150+70" setting in the MTSD dataset following DN-DETR \cite{dn}. b): The performance of CL RT-DETR (CL-DETR using RT-DETR as its baseline), NC CLRT-DETR using NC as fixed class-prototypes (as in \cite{yang2022inducingneuralcollapseimbalanced,ncfscil}) and our method on task 1 in the "150+70" setting of the MTSD dataset.}
    \label{instability}
    \vspace{-3mm}
\end{figure*}

\subsection{Ablations}
We evaluate the effectiveness of HNC in comparison with NC and non-NC in terms of the instability of the Hungarian Matching and the performance over epochs.

\textbf{Class-prototype as GOFs:}

Hungarian matching is a well-known algorithm used in graph matching. DETR was the first algorithm to apply Hungarian matching to object detection, addressing the matching problem between predicted and ground truth objects. On the other hand, a minor modification in the cost matrix can lead to significant changes in the matching outcomes, which in turn may result in conflicting optimization objectives for decoder queries. This leads to instability of the Hungarian Matching in the DETR-based model.

Following Feng et al.~\cite{dn}, we conducted experiments on the instability of the Hungarian Matching of Hier-DETR and CL-RTDETR during 12 epoch training on task 1 of "150+70" setting in the MTSD dataset. Fig.~\ref{instability1} indicates that with the assistance of proxy loss and HNC structure, in the very last decoder layer, it is guided to align to its super-classes and its final predicted class prototype. In other words, HNC ensures that the query retains the characteristics of one-to-many label prediction in the early decoder layers while achieving one-to-one label assignment in the later decoder layers. As a result, the query is more \textbf{"stable"} in the later decoder layers. The impact of this phenomenon on the convergence issues of DETR-based models remains an open question, which we leave for future work.

\textbf{HNC tree Contruction:}

Fig.~\ref{instability2} also shows that applying directly NC as fixed class prototypes as Yang et al.\cite{ncfscil} is not effective. The model just benefits from NC in the first 10 epochs, then performs the same as the models without NC fixed class prototypes. Intuitively, in the very first epoch, the DETR-based model is not able to discriminate the difference between each class's features, so applying NC as a class prototype can help the query's class classification. But, when the model is trained "long" enough, this kind of regularization does not adaptively, especially by forcing each query to align to fixed (randomly initialized) class prototypes in only the last decoder layer. 

In contrast, our method "teaches" the query to align to its super-classes at the initial decoder layers, then gradually makes it allocate to its "final" class prototype by loss proxy.

\section{Conclusions And Limitations}
 In this paper, we have presented a remarkable practical end-to-end Class Incremental Transformer detector, \textbf{Hier-DETR}, with the construction of HNC and loss proxy, which considerably improves the performance of CIOD frameworks on both COCO and MTSD datasets across all continual settings. Our Hier-DETR approach is easy to apply to any kind of DETR-based model in IOD. We have also investigated on the effects of HNC on the instability of the Hungarian Matching for the transformer detection model. For future work, we will explore the impacts of the \textbf{"stability"} of query in the last decoder layers on the model's convergence.

 \textbf{Limitations.} While our approach effectively addresses the strategy of applying both Neural Collapse and the hierarchical relationship of class labels, it does exist few limitations. Although the query feature changes during training, each prototype of GOFs is fixed at the beginning, which can negatively effect on the convergence of aligning query feature to HNC structure. In spite of these challenges, our methods significantly surpasses all previous approaches. Moreover, we believe that, to a certain degree, our framework mitigates this convergence problem by applying class hierarchy at different decoder layers.

\bibliography{ref}

\end{document}